%% file: main.tex
\documentclass[10pt,twocolumn,letterpaper]{article}

\usepackage{cvpr}     

\usepackage{graphicx}
\usepackage{multirow}

\input{preamble}

\definecolor{cvprblue}{rgb}{0.21,0.49,0.74}
\usepackage{hyperref}
\hypersetup{pagebackref,breaklinks,colorlinks,citecolor=cvprblue}

\def\paperID{1250} % *** Enter the Paper ID here
\def\confName{CVPR}
\def\confYear{2024}

\title{Unsupervised Learning of Category-Level 3D Pose from Object-Centric Videos}%Unsupervised Multi-View Object Alignment using Viewpoint Invariant Surface Features}

\author{Leonhard Sommer \textsuperscript{1}
\and
Artur Jesslen \textsuperscript{1}
\and
Eddy Ilg \textsuperscript{2}
\and
Adam Kortylewski \textsuperscript{1,3}
\and
\\
\textsuperscript{1}University of Freiburg \quad \textsuperscript{2}Saarland University \quad \textsuperscript{3}Max Planck Institute for Informatics
}

\begin{document}
\maketitle
\input{sec/0_abstract}    
\input{sec/1_intro}
\input{sec/2_related_work}
\input{sec/3_method}
\input{sec/4_experiments}
\input{sec/5_conclusion}

{
    \small
    \bibliographystyle{ieeenat_fullname}
    \bibliography{main}
}
\clearpage \newpage
\input{sec/6_supplementary}

\end{document}

%% file: preamble.tex
\usepackage[dvipsnames]{xcolor}

%% file: sec/0_abstract.tex
\begin{abstract}
    Category-level 3D pose estimation is a fundamentally important problem in computer vision and robotics, e.g. for embodied agents or to train 3D generative models.
     However, so far methods that estimate the category-level object pose require either large amounts of human annotations, CAD models or input from RGB-D sensors. 
     In contrast, we tackle the problem of learning to estimate the category-level 3D pose only from casually taken object-centric videos without human supervision. 
     We propose a two-step pipeline:  First, we introduce a multi-view alignment procedure that determines canonical camera poses across videos with a novel and robust cyclic distance formulation for geometric and appearance matching using reconstructed coarse meshes and DINOv2 features. 
     In a second step, the canonical poses and reconstructed meshes enable us to train a model for 3D pose estimation from a single image. 
     In particular, our model learns to estimate dense correspondences between images and a prototypical 3D template by predicting, for each pixel in a 2D image, a feature vector of the corresponding vertex in the template mesh.
     We demonstrate that our method outperforms all baselines at the unsupervised alignment of object-centric videos by a large margin and provides faithful and robust predictions in-the-wild.
     %  on the Pascal3D+ and ObjectNet3D datasets
    Our code and data is available at \href{https://github.com/GenIntel/uns-obj-pose3d}{https://github.com/GenIntel/uns-obj-pose3d}.
\end{abstract}

%% file: sec/1_intro.tex
   \begin{figure*}
  \centering
   \includegraphics[width=\linewidth]{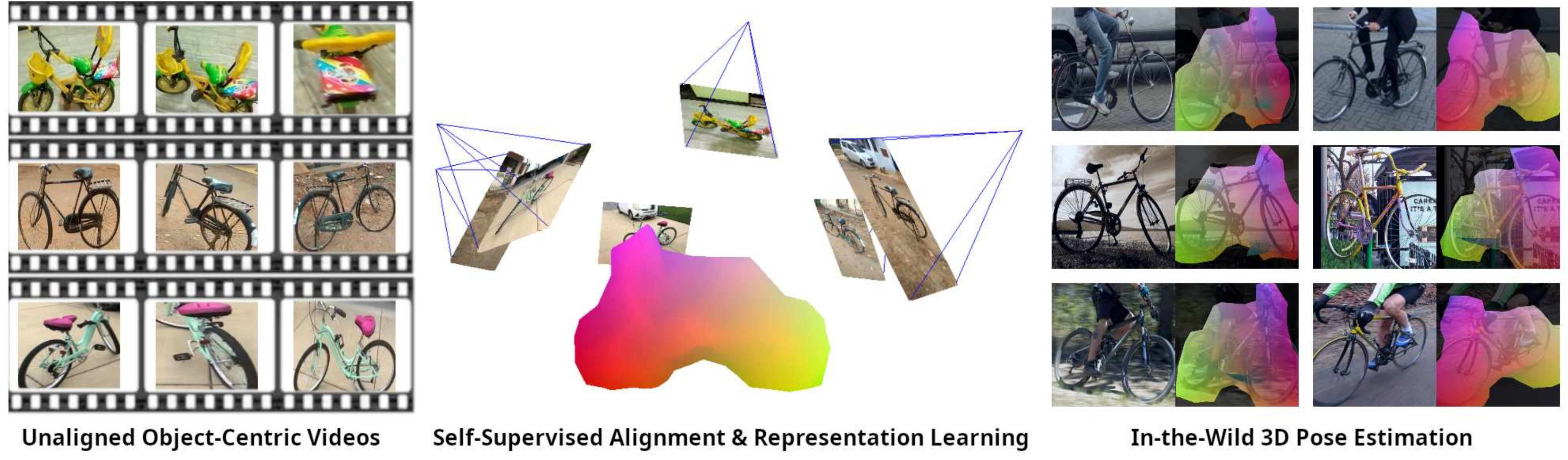}
   \caption{Illustration of our approach for the unsupervised learning of category-level 3D pose. Our method starts from unaligned object-centric videos of an object category (left) and aligns these into a canonical coordinate frame in a self-supervised manner using a prototypical 3D mesh and self-supervised transformer features (center). Using the aligned videos, we train a neural network backbone to predict 2D-3D correspondences from a single image to enable 3D object pose estimation in the wild (right).}
   \label{fig:teaser}
\end{figure*}

\section{Introduction}
\label{sec:introduction}

Category-level object pose estimation is a fundamentally important task in computer vision and robotics with a multitude of
real-world applications, e.g. for training 3D generative models on real data and for robots that need to grasp and manipulate objects. 
However, defining and determining the pose of an object is a task that is far from easy. 

%Current approaches to category-level 3D pose estimation 
Current approaches achieve high performance, but they require large amounts
of annotated training data to generalize successfully \cite{zhou2018starmap,akizuki2021asm,wang2021nemo,wang2019normalized,xiao2021posecontrast}, or additional inputs during inference, such as CAD Models~\cite{grabner20183d,labbe2022megapose}, 3D Shapes~\cite{xiao2019pose,chen2021sgpa,zhang2022self} or \mbox{RGB-D}~\cite{gupta2015inferring}. 
However, all of these are either time-consuming to obtain or not available in practice at all. This motivates the development of methods for learning category-level 3D pose estimators in a fully unsupervised fashion.
While doing so from images in the wild seems infeasible, object-centric video data~\cite{reizenstein2021co3d} offers a more accessible alternative. Such videos can be easily captured using consumer-grade cameras and makes it possible to leverage coarse 3D reconstructions during training, providing a practical and cost-effective method for collecting data. 
Therefore, we propose the new task of learning a single-image category-level 3D pose estimator from casually captured object-centric videos without any human labels or other supervision. In practice, we leverage CO3D~\cite{reizenstein2021co3d} as training data and show that our proposed model is able to generalize and predict accurate poses in the wild for Pascal3D+ \cite{xiang2014beyond} and ObjectNet3D \cite{xiang2016objectnet3d}. 
We address the challenging task of learning category-level 3D pose in an unsupervised fashion from casually captured object-centric videos. In particular, we propose a two-step pipeline (Figure \ref{fig:teaser}). 
The first step extracts DINOv2~\cite{oquab2023dinov2} features from the images and reconstructs a coarse 3D mesh from the video with off-the-shelf methods~\cite{kirkpatrick1983shape,garland1997surface}. 
Building on this input, we introduce a novel 3D alignment procedure, where a key contribution is a novel 3D cyclical distance in terms of geometry and appearance that enables the robust alignment of shape reconstructions even under severe noise and variations in the object topology. 
As a result, we can align all objects from the object-centric training videos into a canonical coordinate frame without supervision. 
In a second step, we leverage the canonical poses and 3D meshes obtained from the first step to train a category-level neural mesh \cite{neverova2020continuous,wang2021nemo,iwase2021repose,ze2022category} in an unsupervised manner.
In particular, we represent objects using a prototypical 3D mesh with surface features to capture the geometry and neural appearance of an object category 
and train a neural network backbone to predict, for each pixel in a 2D image, a feature vector of the corresponding vertex in the template mesh.
Finally, the object 3D pose is solved using a pose fitting algorithm based on the estimated correspondence pairs.
We demonstrate that our method outperforms all baselines by a large margin at the unsupervised alignment of object-centric videos on the CO3D \cite{reizenstein2021co3d} dataset.
Moreover, our model provides faithful and robust predictions in-the-wild on Pascal3D+ \cite{xiang2014beyond} and ObjectNet3D \cite{xiang2016objectnet3d} despite being trained from raw object-centric videos only. 

%% file: sec/2_related_work.tex
\section{Related Work}
\label{sec:relatedwork}
% q1: is canonicalized camera poses a valid term?

\textbf{Supervised Category-Level 3D Pose Estimation.}  Traditional methods to determine object poses were to label keypoints in images and train supervised methods to predict them~\cite{tulsiani2015viewpoints,zhou2018starmap}, or to use only pose labels and directly predict them by casting the pose estimation problem as bin classification by discretizing the pose space~\cite{su2015render}. More recent methods utilize 3D meshes of objects or object categories. NeMo~\cite{wang2021nemo} uses 3D meshes with neural features that are rendered-and-compared to feature maps from a CNN to obtain pose estimates. \cite{kouros2022category}~predicts a single embedding vector per image and shows that superior performance can be obtained by simply retrieving the closest training sample. 
In contrast to the above methods, our method is unsupervised. 

\textbf{Few-Shot and Zero-Shot Pose Prediction.} \cite{xiao2021posecontrast} proposes to train a supervised pose estimator across many categories and shows that their approach can generalize well to similar but unseen objects.  Many works implement zero shot pose estimation by conditioning the model on the 3D shape of the unseen object~\cite{xiao2019pose,chen2021sgpa,zhang2022self,xiao2022few} or by leveraging renderings of CAD models~\cite{grabner20183d,labbe2022megapose}. Other methods use few-shot learning~\cite{tseng2019few} or zero-shot learning with DINOv2~\cite{oquab2023dinov2}. In contrast to the above, our method does not require any annotated dataset, 3D shapes or CAD models. 

\textbf{Pose Alignment.}
The work from Goodwin et al.~\cite{goodwin2022zsp} (ZSP) is most close to the first step of our method, as it aligns the poses of two object-centric videos in a fully unsupervised fashion. Similar to our work, they use DINO~\cite{caron2021dino} to obtain semantic correspondences.
They perform first a coarse alignment by matching one image from the source video to one of many images from the reference video. Then they leverage cyclical distances to select few promising correspondences in the two images, and finally leverage respective depth maps to align both images using least squares.  
Goodwin et al. extend their work in ~\cite{goodwin2023yolo} (UCD+) to match many images from the source video to many of the reference video. By finding a consensus over these many to many alignments with a single transformation they demonstrate improved performance. The first step of our work is similar to these works by that it also leverages DINO features and cyclical distances. However, our work adds a geometric distance to perform the alignment directly in 3D and introduces weighted correspondences to enable the necessary robust regression of the SE3 transformations from noisy and inaccurate geometries, which leads to significant improvements in the alignment accuracy. Note also that these previous approaches used RGB-D inputs, while our method works on images directly.

\textbf{Surface Embeddings and Neural Mesh Models.}
Recent work uses known poses and approximate object geometries to learn features in 3D space to uniquely identify parts of objects.  \cite{neverova2020cse} first used known mesh templates of deformable objects to train a network that predicts surface embeddings from the images. NeMo~\cite{wang2021nemo} presents a generative model trained with contrastive learning. 
\cite{wang2022voge} presents an extension through replacing vertices by Gaussian ellipsoids and using volume rendering. Similar to our work, many recent works leverage pre-trained vision 
tranformers DINO~\cite{caron2021dino} and DINOv2~\cite{oquab2023dinov2} to unproject image features onto  depth maps~\cite{goodwin2022zsp,goodwin2023yolo,zhang2022self}. 
In contrast to \cite{neverova2020cse,wang2021nemo,wang2022voge}, our approach does not require any pose annotation, while also going beyond \cite{goodwin2022zsp,goodwin2023yolo} by enabling 3D pose estimation in the wild from a single image, and not requiring RGB-D images and CAD models as input \cite{zhang2022self}.

%% file: sec/3_method.tex
\section{Method}
\label{sec:method}

\begin{figure}
  \centering
  \includegraphics[width=\linewidth]{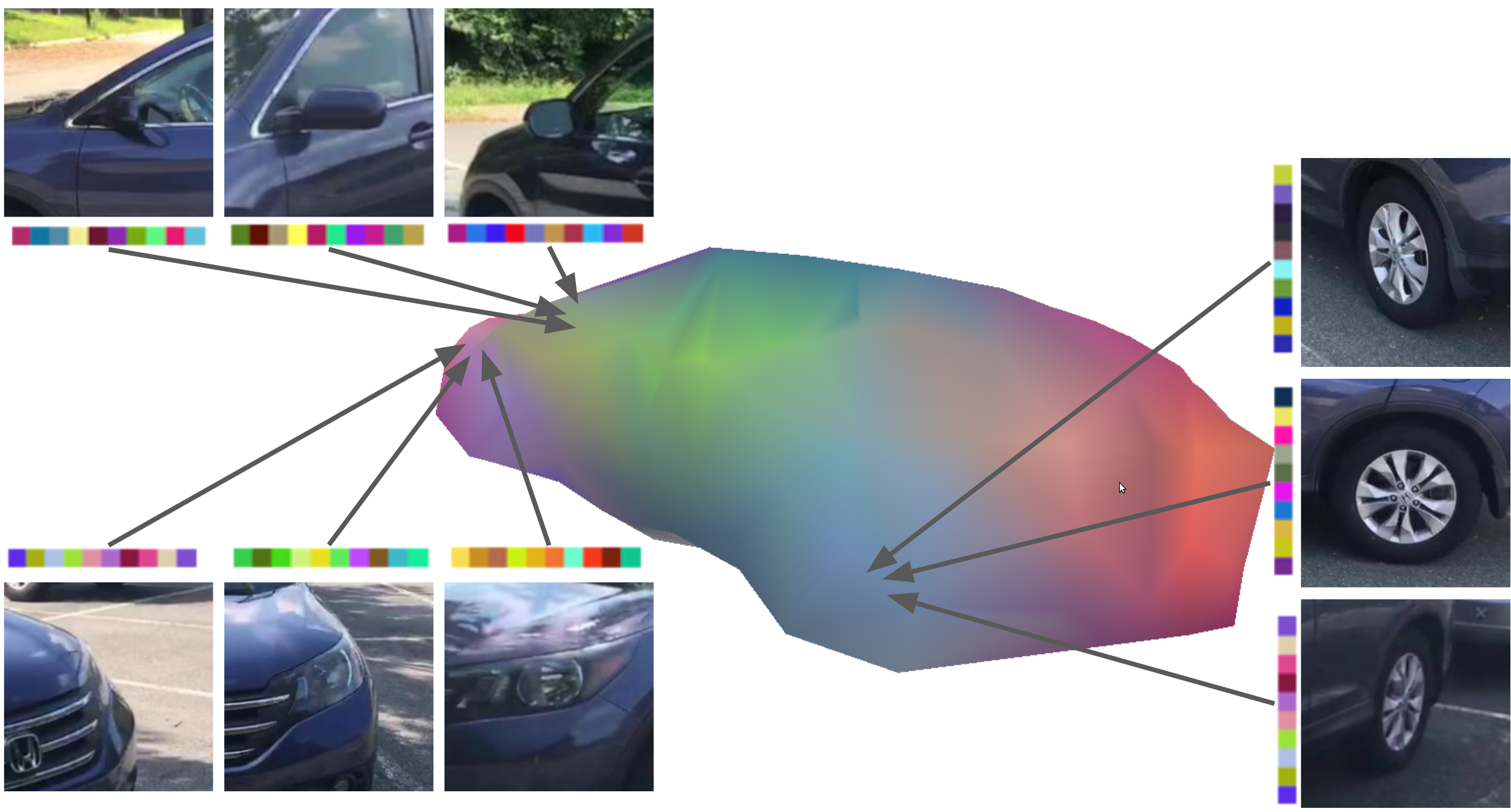} % ,height=7cm
   \caption{Illustration of a neural mesh. It consists of a 3D mesh with one or several neural features per vertex. These features encode patch-level features from a feature extractor. In our unsupervised alignment method, we capture the viewpoint-dependent features from DINOv2. For pose estimation, we learn a category-level neural mesh with viewpoint-invariant features.}
   \label{fig:pipeline}
\end{figure}

\newcommand{\argmin}{\arg\!\min}
\newcommand{\verts}{\mathrm{V}}
\newcommand{\refverts}{\bar{\verts}}
\newcommand{\srcverts}{\verts}
\newcommand{\Nemo}{Surface Features}
\newcommand{\Nemos}{Surface Features}
\newcommand{\nemo}{surface features}
\newcommand{\nemos}{surface features}

\newcommand{\meshsymbol}{\mathrm{M}}
\newcommand{\meshword}{mesh}
\newcommand{\Meshword}{Mesh}

\newcommand{\distcycle}{\mathrm{d}_{\mathrm{cycle}}}
\newcommand{\distrange}{\mathrm{d}_{\mathrm{range}}}
\newcommand{\distapp}{\mathcal{D}_{\mathrm{app}}}
\newcommand{\distappvertices}{\mathrm{d}_{\mathrm{app}}}
% \mathrm{d}_{\mathrm{app.}}
\newcommand{\distgeo}{\mathcal{D}_{\mathrm{geo}}}
\newcommand{\distchamfer}{\mathcal{D}_{\mathrm{chamfer}}}
\newcommand{\distnemo}{\mathcal{D}}

\newcommand{\good}{\rho}
\newcommand{\weight}{\sigma}
\newcommand{\tempcyclic}{\tau}

\newcommand{\vertexsymbol}{\mathrm{v}}

\newcommand{\vi}{\vertexsymbol_i}

\newcommand{\vj}{\vertexsymbol_j}
\newcommand{\fj}{\feat_j}
\newcommand{\prob}{\mathrm{p}}

\newcommand{\vv}{\vertexsymbol}
\newcommand{\vcyclce}{(\vv^{\mathrm{nn,f}})_{}^{\mathrm{nn,f}}}
\newcommand{\vnnfeat}{\psi{}} %  \vertexsymbol^{\mathrm{nn,f}}
\newcommand{\vnneuc}{\chi{}}  %  \vertexsymbol^{\mathrm{nn,x}}
\newcommand{\vinnfeat}{\vertexsymbol_i^{\mathrm{nn,f}}}
\newcommand{\vinneuc}{\vertexsymbol_i^{\mathrm{nn,x}}}
\newcommand{\vjnnfeat}{\vertexsymbol_j^{\mathrm{nn,f}}}
\newcommand{\vjnneuc}{\vertexsymbol_j^{\mathrm{nn,x}}}

\newcommand{\vertexsrcsymbol}{\vertexsymbol_{\mathrm{src}}}
\newcommand{\vertexrefsymbol}{\vertexsymbol_{\mathrm{ref}}}
\newcommand{\featbgsymbol}{\mathrm{b}}

\newcommand{\vertexword}{vertex}
\newcommand{\Vertexword}{Vertex}

\newcommand{\imagesymbol}{\mathrm{I}}
\newcommand{\imageword}{Image}

\newcommand{\pixelsymbol}{\mathrm{p}}
\newcommand{\pixelword}{Pixel}

\newcommand{\encodersymbol}{\phi}
\newcommand{\encoderword}{Image Encoder}

\newcommand{\feats}{\mathrm{F}}
\newcommand{\feat}{\mathrm{f}}
\newcommand{\featword}{Feature}
\newcommand{\featsword}{Features}

\newcommand{\reffeats}{\bar{\feats}}
\newcommand{\srcfeats}{\feats}

\newcommand{\locpointsymbol}{\mathrm{x}}
\newcommand{\locpointspacesymbol}{\mathbb{R}^3}

\newcommand{\locpixelsymbol}{\mathrm{u}}
\newcommand{\locpixelspacesymbol}{\mathbb{R}^2}

\newcommand{\featspacesymbol}{\mathbb{R}^D}
\newcommand{\featspacemdsymbol}{\mathbb{R}^{D}}

\newcommand{\tformsymbol}{\mathrm{T}}
\newcommand{\tform}{\mathrm{T}} %_{\mathrm{src}}^{\mathrm{ref}}}

\newcommand{\camtformsymbol}{\mathrm{T}_{\mathrm{obj}}^{\mathrm{cam}}}

\newcommand{\projectionsymbol}{\pi}

% TODO: Explain how we have 
% 3D Representation: Triangular Surface Mesh
In this section, we describe our approach for learning category-level 3D pose estimation without supervision from object-centric videos. Our method proceeds in a two-step approach. First, we align object instances across videos in an unsupervised manner to bring them into a canonical reference frame (Section~\ref{sec:method_alignment}). 
Given the aligned videos, our model learns to establish dense correspondences between images and a reconstructed 3D template mesh by predicting a feature vector for each pixel in a 2D image that corresponds to a visible vertex in the template
mesh (Section~\ref{sec:method_learn_nemo}). 
Finally, we describe how our model can efficiently estimate object poses from in-the-wild data using the predicted correspondences via render-and-compare.

% Surface Features
\subsection{Meshes with \Nemos}
In both steps of our approach, the video alignment and the representation learning, we represent objects as \textit{neural meshes}, i.e. meshes with \nemos{} to capture the geometry and appearance of an object instance or category \cite{iwase2021repose,wang2021nemo,ze2022category,neverova2020continuous}. 
In particular, the geometric representation is a triangular mesh, where we denote the set of vertices as $\verts{} = \{\vv_i \in \mathbb{R}^3\}_{i=1}^{|V|}$. 
The appearance is represented by storing  one or multiple appearance features $\feats=\{\{\feat_{i}^\mathrm{k}\in\mathbb{R}^D\}_{k=1}^{|F|}\}_{i=1}^{|V|}$ at each mesh vertex.
Together, the geometry and appearance define a neural mesh as $S=\{V,F\}$.

\subsection{Self-supervised Alignment of Objects}
\label{sec:method_alignment}
Our goal is to align the camera poses of multiple object-centric videos into a common coordinate frame. 
To achieve this, we represent each object-centric video as a neural mesh with self-supervised surface features. 
In particular, we utilize off-the-shelf structure-from motion \cite{schonberger2016colmap} to obtain a coarse object shape reconstruction for each video.
Note that the reconstructed shapes cover the whole object, as the object-centric videos move in a full circle around the object.
We post-process the reconstructed point cloud to clean it and generate a watertight mesh, for which we provide details in the supplementary material. 
Subsequently, we project the reconstructed coarse meshes into the feature map $\Psi(I)$ that is obtained from a self-supervised transformer backbone \cite{oquab2023dinov2}. We collect from every video a set of feature vectors for every vertex $v_i$ that describe the appearance of a local patch (Figure \ref{fig:pipeline}).
Thus the number of features per vertex depends on the number of images in which the vertex is visible. 
As feature extractor $\Psi$, we use a self-supervised vision transformer \cite{oquab2023dinov2} which has shown emerging correspondence matching abilities.

% Alignment of Surface Features
% Optimization Problem
\textbf{Finding geometric and appearance correspondences.} Given the mesh vertices of the source object instance $\srcverts$ and the corresponding aggregated features $\srcfeats$, we aim to align them to the reference counterparts $\refverts$ and $\reffeats$. 
In practice, we select a reference video at random from the set of all available videos and align the remaining videos to the reference.
More precisely, we aim to optimize the transformation $\tform$, which is composed of rotation, translation and scale, under which the transformed source vertices and corresponding features yield the minimal distance to the reference counterparts with respect to geometry as well as appearance. Formally, our optimization problem optimizes
\begin{equation}
    \min_{\tform{}} \distnemo(S, \bar{S}, \tform){=}\distgeo(\srcverts, \refverts)+\distapp(S, \bar{S}),    
\end{equation}
where $\distnemo$ is the similarity between two videos given a transformation $T$, which combines a geometric distance between the mesh geometries $\distgeo(\srcverts, \refverts)$ and an appearance distance $\distapp(S, \bar{S})$ between surface features.

% Geometric Distance
Assuming that the object instances shapes contain a negligible variance and no symmetries, a suitable geometric distance is the Chamfer Distance defined as
% Chamfer Distance
%\resizebox{\linewidth}{!}{
\begin{equation}
\begin{aligned}\distgeo(\srcverts, \refverts) = 
     \sum_{\vv_i \in \srcverts \cup \refverts} || \vv_i - \vv_{\vnneuc(\vv_i)}||_2,
\end{aligned}
\end{equation}
where $\vnneuc(\vv_i)$ is the vertex index of the Euclidean nearest neighbor of vertex $\vv_i$ in the respective other set of vertices
% nearest neighbor euclidean space
\begin{equation}
    \vnneuc(\vv_i) = 
    \begin{cases}  
         \underset{{j \in 1 ... |\refverts{}|}}{\argmin} || \tform \vi - \vj||_2  \,,& \text{for } \vi \in \srcverts{}  \\
          \underset{{j \in 1 ... |\srcverts{}|}}{\argmin}  || \tform \vj - \vi||_2  \,, &  \text{for } \vi \in \refverts{}.  
    \end{cases}
\end{equation}

% Appearance Distance
However, as the 3D object is rather coarse and noisy, and the alignment can be ambiguous due to symmetries or shape differences among objects, we optimize each vertex to also be geometrically close to its nearest neighbor in feature space using the appearance distance
\begin{equation}
\begin{aligned}\distapp(S, \bar{S}) = 
     \sum_{\vv_i \in \srcverts \cup \refverts} || \vv_i - \vv_{\vnnfeat(\vv_i,\feat_i)}||_2,
\end{aligned}
\end{equation}
with the nearest neighbor in feature space $\vnnfeat(\vv)$ defined as 
% nearest neighbor feature space
\begin{equation}
    \vnnfeat(\vi, \feat_i) = 
    \begin{cases}  
        \underset{{j \in 1 ... |\refverts{}|}}{\argmin} \min_{k,l} || \feat_{j}^k - \feat_{i}^l|| \,, & \text{for } \vv_i \in \srcverts \\
        \underset{{j \in 1 ... |\srcverts{}|}}{\argmin} \min_{k,l} || \feat_{j}^k - \feat_{i}^l|| \,, & \text{for } \vv_i \in \refverts.
    \end{cases}
\end{equation}

%Each video provides multiple features per vertex from different viewpoints. 
As the self-supervised vertex features are view-dependent, the appearance distance computes the minimum feature distance across all views to select the nearest neighbor.

% Weighting Correspondences
\textbf{Weighting Correspondences.} An open challenge for the alignment of casually captured object-centric videos is that the estimated correspondence pairs between videos can be unreliable. For example, due to errors in the shape reconstruction or significant topology changes among different object instances, such as one bicycle having support wheels whereas the other does not. 
In these cases the correspondences in the geometry and feature space are ill-defined which leads to unreliable correspondence estimates. 
To account for such unreliable correspondences, we introduce a weight factor for each correspondence pair that estimates its quality.
At the core of the correspondence weighting, we introduce a 3D cyclical distance among the vertices of two neural meshes that is inspired by 2D cyclical distances \cite{goodwin2022zsp} for correspondence estimation, and is defined as
% cyccle distance
\begin{equation}
    \distcycle{}(\vi, \feat_i) = || \vi - \vv_{\vnnfeat(\vv_j,\feat_j)} ||_2,\hspace{.1cm}\text{with}\hspace{.1cm}j={\vnnfeat(\vi, \feat_i)}. 
\end{equation}
The nested structure of our 3D cyclical distance first computes $j$ as the index of the nearest neighbor of vertex $\vv_i$ in the feature space, and in turn computes the nearest neighbor of $\vv_j$ as $\vv_{\vnnfeat(\vv_j,\feat_j)}$.
Notably, $\distcycle{}(\vi, \feat_i)=0$ if the nearest neighbour maps back to the original vertex  $\vv_{\vnnfeat(\vv_j,\feat_j)}=\vi$ and hence the correspondence is reliable.
% quality measure
Building on this 3D cyclical distance, we define the validity criteria for each pair of vertices as the sum of cyclical distances of the correspondence pair
\begin{equation}
\begin{aligned}
    \good(\vi\,, \feat_i\,, \vj\,, \feat_j) =  - \frac{\distcycle{}(\vi, \feat_i) + \distcycle{}(\vj, \feat_j)}{2 \tempcyclic (\mathrm{D}(\srcverts) + \mathrm{D}(\refverts))},
\end{aligned}
\end{equation}
where $\mathrm{D}(\cdot)$ is the diameter of a neural mesh given as $\mathrm{D}(\verts) = \max_{\vi\,, \vj \in \verts} || \vi - \vj  ||_2$. %For a large cyclical distance the validity score decreases.  

% softmax distribution
To obtain the final weight factor for a correspondence pair we use the softmax normalization $\weight_\tau(i,j)=\texttt{Softmax}_\tau(\good(\vi\,, \feat_i \,, \vj\,, \feat_j))$,  across all feature and gemoetric correspondences. 
For the softmax normalization, we introduce the temperature $\tau$, which enables us to steer between taking into account fewer high quality correspondences or more low quality ones (see Section \ref{sec:exp:abl}).
Together with the weighting we formulate the weighted geometric distance as
% weighted geo distance
\begin{equation}
\distgeo^*(S, \bar{S}) {=}  
     \sum_{\vv_i \in \srcverts \cup \refverts} \weight(i,\vnneuc(\vv_i)) \hspace{.1cm} || \vv_i - \vv_{\vnneuc(\vv_i)}||_2,
\end{equation}
and likewise the weighted appearance distance as
% weighted appearance distance
\begin{equation}
\distapp^*(S, \bar{S}) =  
     \sum_{\vv_i \in \srcverts \cup \refverts} \weight(i,\vnnfeat(\vv_i)) \hspace{.1cm} || \vv_i - \vv_{\vnnfeat(\vv_i)}||_2.
\end{equation}
% total distance
Our final distance measure to compare two neural meshes is computed as 
%\resizebox{\linewidth}{!}{
\begin{equation}    
\begin{aligned}
    \mathcal{L}(S, \bar{S}) =
    (1 - \alpha) \distgeo^*(S, \bar{S})  + \alpha  \distapp^*(S, \bar{S}) .
\end{aligned}
\end{equation}
%}

To find an approximately optimal solution, we use a RANSAC strategy, where we randomly choose four vertices on the source surface mesh. Together with their nearest neighbors in the feature space on the reference surface mesh, we estimate a single transformation using the Umeyama method \cite{umeyama1991least}.
\subsection{3D Pose Estimation In-the-Wild}
\label{sec:method_learn_nemo}
Our goal is to perform 3D pose estimation in in-the-wild images.
To achieve this, we generalize our approach from the multi-view setting used to align object-centric videos, towards 3D pose inference from a single image.
Our model uses a feature extractor $\Psi_w(I) = F \in \mathbb{R}^{D \times H\times W}$ to obtain image features from input image $I$, where $w$ denotes the parameters of the backbone. The backbone output is a feature map $F$ with feature vectors $f_i \in \mathbb{R}^{D}$ at positions $i$ on a 2D lattice.
For training, we use the aligned object-centric videos (Section \ref{sec:method_alignment}) to train the weights $w$ of the feature extractor $\Psi_w$ such that it predicts dense correspondences between image pixels and the 3D neural mesh template. 
Specifically, 
we relate the features of an image $\Psi_w(I)$ extracted by a backbone feature extractor to the vertex and background features by Von-Mises-Fisher (vMF) probability distributions \cite{kortylewski2020compositional}.
In particular, we model the likelihood of generating the feature at an image pixel $f_i$ from corresponding vertex feature $f_r$ as $P(f_i | f_r) = c_p(\kappa) e^{\kappa f_i \cdot f_r}$, where  $f_r$ is the mean of each vMF kernel, $\kappa$ is the corresponding concentration parameter, and $c_p$ is the normalization constant ($\lVert f_i \lVert = 1, \lVert f_r \lVert = 1$).
We also model the likelihood of generating the feature $f_i$ from background feature as $P(f_i | \beta) = c_p(\kappa) e^{\kappa f_i \cdot \beta} $ for $\beta \in \mathbb{R}^{D}$. 

When learning the models, as described next, we will learn the vertex features $\{f_r\}$, the background feature $\beta \in \mathbb{R}^{D}$, and the parameters $w$ of the neural network backbone $\Psi_w$. We emphasize that our model requires that the backbone must be able to extract features that are invariant to the viewpoint of the object to ensure that $f_i \cdot f_r$ is large irrespective of the viewpoint.

\textbf{Learning viewpoint-invariant vertex features.} For training our model, we use the visible vertex features and their corresponding image features $\mathcal{P} = \{(f_r, f_i)\}$. Further, we use image features randomly sampled from the background $\mathcal{B} = \{f_i \}$. As optimization objective, we use the cross-entropy loss
\begin{align}
    \mathcal{L}_{train} =  
    - \sum_{(f_r, f_i) \in \mathcal{P}} \log\left(\frac{P(f_i|f_r)}{\sum{P(f_i|f_r}) + \sum{P(f_i|\beta) }} \right) \nonumber \\
    - \sum_{f_i \in \mathcal{B}} \log\left(\frac{P(f_i | \beta)}{\sum{P(f_i|f_r}) + \sum{P(f_i|\beta) }}\right). 
\end{align}

\textbf{3D pose inference.} We use the mesh with the vertex features $\{f_r\}$, the background feature $\beta$ and the trained backbone $\Psi_w$ to estimate the camera pose $\alpha$ via render-and-compare. At each optimization step, we render a feature map $\{ \bar{f}_i(\alpha) \}$ under pose $\alpha$ and compare it with the encoder's feature map $F = \{ f_i \}$. Determined by the rendering, each feature map consists of foreground features $F_{front}$ and background features $F_{back} = F \setminus F_{front}$. Thereupon, we maximize the joint likelihood for all image features under the assumption of independence, given as
\begin{align}
    P(F | \alpha, \{ f_r \}, \beta) = \nonumber \\ 
    \prod_{f_i \in F_{front}} \max_{f' \in \{ \bar{f}_i(\alpha), \beta \}}P(f_i | f') 
    \prod_{f_i \in F_{back}} P(f_i | \beta).
\label{eq:reconstruction_loss}
\end{align}
Note, by allowing foreground image features to be generated by the background feature, we also account for clutter.

We estimate the pose by first finding the best initialization of the object pose $\alpha$ by computing the joint likelihood (Eq.\ref{eq:reconstruction_loss}) for a set of pre-defined poses via template matching and choosing the one with the highest likelihood. 
Subsequently, we iteratively update our initial pose using a differentiable renderer to obtain the final pose prediction $\hat{\alpha}$.

%% file: sec/4_experiments.tex
\begin{table*}
\centering
\begin{tabular}{l|rrrrrr|r|r}
    
    \hline
     & \multicolumn{6}{c}{Per Category} & \multicolumn{2}{|c}{All Categories (20)}  \\
        &   backpack &    car &   chair &   keyboard &   laptop &   motorcycle &   Acc. $30^\circ$ & Acc. $15^\circ$  \\
    \hline
    TEASER++ \cite{yang2020teaser}  & 1.0 & 5.0 & 9.0 & 8.0 & 6.0 & 1.0  & 3.8  & 1.1 \\ 
    %  & 1.1 & 0.3 & 126.4 (pi/12, pi/24, median)
    ZSP \cite{goodwin2022zsp}   & 44.0 & 65.0 & 47.0 & 69.0 & 85.0 & 85.0 & 49.4 & 28.4  \\
    %  & 28.35 & 9.6 & 47.7
    UCD+ \cite{goodwin2023yolo} & 67.0 & 86.0 & 76.0 & 76.0 & \textbf{100.0} & \textbf{100.0} & 69.8 & 54.6  \\
    % & \textbf{54.6} & \textbf{34.5} & \textbf{21.4}   
    \hline
    Ours       &      \textbf{85.6} & \textbf{100.0} &   \textbf{98.9} &      \textbf{82.2} &   \textbf{100.0} &       \textbf{100.0} &       \textbf{77.0} & \textbf{61.6} \\
             &      $\pm$ 10.5 &   $\pm$ 0.0 & $\pm$   3.5 &     $\pm$  10.7 &   $\pm$  0.0 &      $\pm$    0.0 &    $\pm$   11.9 & $\pm$ 15.9 \\
    \hline
    \end{tabular}
\caption{Unsupervised alignment evaluation on the CO3D dataset across 20 categories. The reported metric is the $30^\circ$ accuracy if not stated otherwise. The mean is computed across all 20 categories. We see that our method substantially outperforms the state of the art.}
\label{tab:res_zsp_20}
\end{table*}

\begin{figure*}
    \centering
    \includegraphics[width=\linewidth]{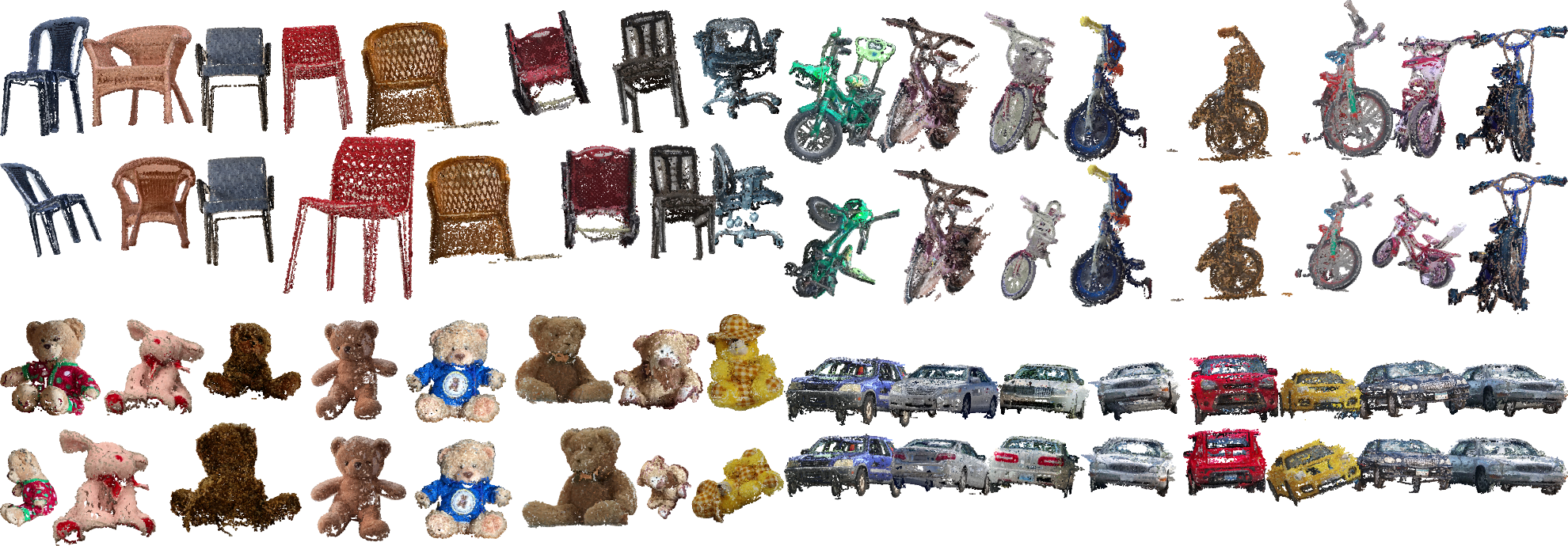} % ,height=7.5cm
    \caption{Qualitative comparison of two unsupervised alignment methods. The first row shows the alignment of our proposed method. The second row shows the alignment using ZSP \cite{goodwin2022zsp}. For both methods we use the 5th object instance from left as reference. We see that our proposed method is more accurate compared with ZSP. Especially for cars ZSP often confuses back and front. 
    }
    \label{fig:experiment_alignment}
\end{figure*}
% ZSP COMPARISON

\section{Experiments}
\label{sec:experiments}
In this section, we discuss our experimental setup (Section \ref{sec:exp:setup}), present baselines and results for unsupervised alignment of object-centric videos (Section \ref{sec:exp:alignment}) and 3D pose estimation in-the-wild (Section \ref{sec:exp:wild}). Additionally, we perform ablations of key model components in Section \ref{sec:exp:abl}.

\subsection{Experimental Setup}\label{sec:exp:setup}
% what do we want to evaluate
\textbf{Dataset for alignment.} To evaluate the unsupervised alignment of object-centric videos, we use the recently released Common Objects in 3D (CO3D) dataset \cite{reizenstein2021co3d} that provides images of multiple object categories, with a large amount of intra-category instance variation, and with varied object viewpoints. It contains 1.5 million frames, capturing objects from $50$ categories, across nearly 19k scenes. For each object instance, CO3D provides approximately $100-200$ frames promising a $360^\circ$ viewpoint sweep with handheld cameras. CO3D supplements these videos with relative camera poses and estimated object point clouds using Structure-from-Motion \cite{schonberger2016colmap}. 

We find that the unfiltered videos of CO3D are not ideal for our purpose. In particular, we find that videos with little viewpoint variation lead to inferior structure-from-motion results. Also, videos that are not focusing on the object's center in 3D or are taken too close to it, contain little information for correspondence learning. Therefore, we filter the videos accordingly, targeting 50 videos per category. For multiple categories we end up with less than 50 videos namely, "remote" 17, "mouse" 15, "tv" 16, "toilet" 7, "toybus" 41, "hairdryer" 28, "couch" 49, and "cellphone" 23. With our simple filters, we end up aiming for 50 videos per category. More precise details for the filtering procedure are appended in the supplementary. 
As labels, we use the ground truth pose annotations provided by ZSP \cite{goodwin2022zsp}, that cover ten object instances of twenty different categories. 
\textbf{Datasets for 3D pose estimation in-the-wild.} We evaluate on two common datasets PASCAL3D+ \cite{xiang2014pascal3d} and ObjectNet3D \cite{xiang2016objectnet3d}. While PASCAL3D+ provides poses for the 12 rigid classes of PASCAL VOC 2012, ObjectNet3D covers pose annotations for over 100 categories. The object-centric video dataset CO3D covers 50 categories from the MS-COCO \cite{lin2014coco} dataset. We find 23 common categories across ObjectNet3D and CO3D, even tolerating the gap between a toybus in CO3D and a real one in PASCAL3D+ and ObjectNet3D. We believe that this non-neglegible gap could be bridged by exploiting the multiple viewpoint knowledge of the same object instance. Overall we validate on PASCAL3D+ with 6233 images, using the same validation set as \cite{wang2020robust}, and on ObjectNet3D on 12039 images. Following \cite{zhou2018starmap}, we center all objects.
\textbf{Implementation details.}
In our alignment step, we use $\tau=100$ and $\alpha=0.2$. Further, we leverage as self-supervised ViT the publicly available small version of DINOv2 \cite{oquab2023dinov2} with 21M parameters and a patch size of 14. At the input we use a resolution 448x448 ending up with a 32x32 feature map, where each feature yields 384 dimensions. In our second step, we use the same ViT as backbone and freeze its parameters. Further, we add on top three ResNet blocks with an upsampling step preceding the final block. Ending up with a 64x64 feature map, where each feature has 128 dimensions. We optimize the cross-entropy loss for 10 epochs with Adam \cite{kingma2014adam}. In one epoch, we make use of all filtered videos. The training for each category-level representation takes less than an hour on a single NVIDIA GeForce RTX 2080.

We note that the quality of our alignment method and the subsequent representation learning can vary depending on the chosen reference video. Therefore, we randomly choose five reference videos per category and report the mean performance and the standard deviation across all results.

% evaluation metric
We report the $30^\circ$ accuracy for pose estimation where the angle error for an estimated rotation $\mathrm{R}_{pred}$ and a ground truth rotation $\mathrm{R}_{pred}$ is given as
\begin{equation}
        \Delta(\mathrm{R}_{pred}, \mathrm{R}_{gt}) = \arccos \left( \frac{1}{2}\mathrm{tr}\left( \mathrm{R}_{pred}^{\mathrm{T}} \mathrm{R}_{gt} \right) -1 \right).
\end{equation}
We note that for the current state of unsupervised pose estimation, achieving $30^\circ$ precision remains unsolved. 

\subsection{Unsupervised Alignment}
\label{sec:exp:alignment}

We follow the evaluation protocol of ZSP \cite{goodwin2022zsp} and measure the alignment of one object instance to the nine remaining ones of the same category that are labelled. Additionally, we report the standard deviation across the chosen reference object instances. The quantitative results in Table \ref{tab:res_zsp_20} show that our proposed method significantly improves the state of the art by $7.2\%$ from $69.8\%$ to $77.0\%$. Our alignment algorithm can more efficiently use the video frames compared to ZSP \cite{goodwin2022zsp}, which only compares a single RGB-D frame from the source video with many RGB-D frames of the reference video. We note that ZSP uses DINOv1 features in contrast to our method, which uses DINOv2. Therefore, we provide an ablation of our method with respect to different feature extractors in the supplementary.
One reason for our model to outperform UCD+ \cite{goodwin2023yolo}, an extension of ZSP, is likely that our optimization does exploit the object geometry extensively, whereas others are using it only for refinement. A qualitative comparison of our method against ZSP is depicted in Figure \ref{fig:experiment_alignment}. It shows that our alignments are highly accurate despite a large variability in the object instances. We note that at the time of writing, there is no source code publicly available to compare with UCD+.
\subsection{In-the-Wild 3D Pose Estimation}
\label{sec:exp:wild}

\begin{figure*}
    \centering
    \includegraphics[width=\linewidth]{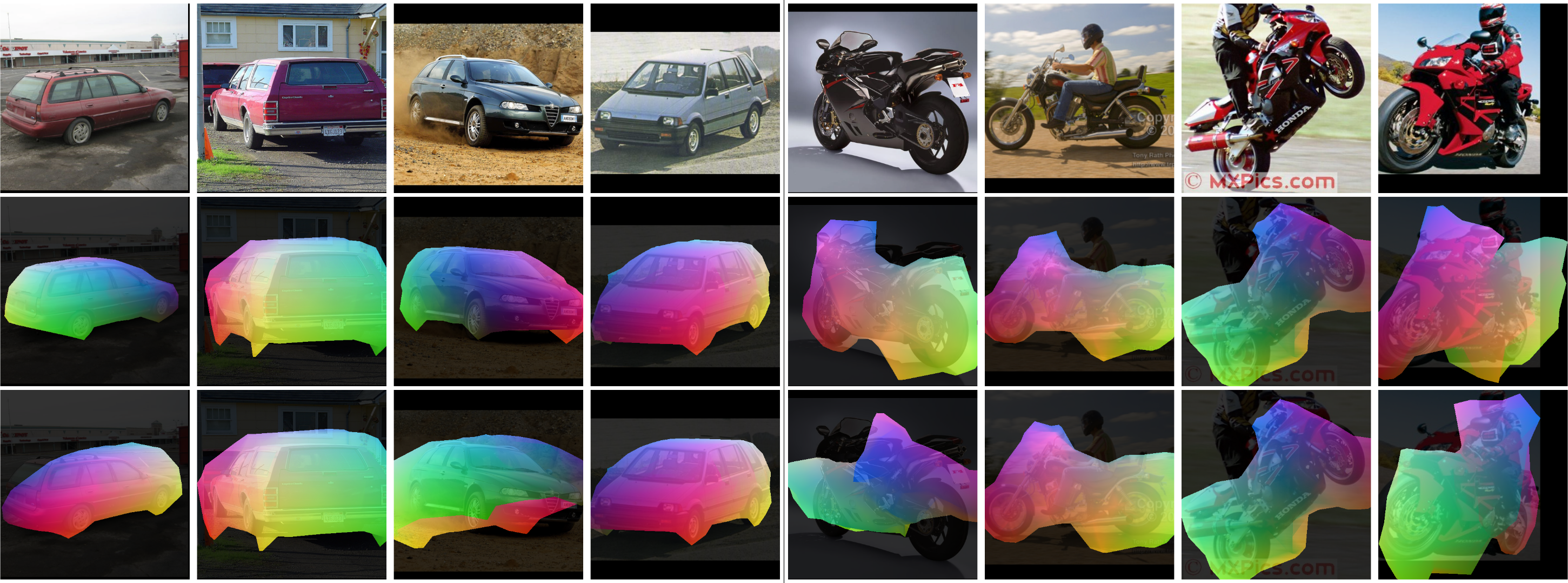}
    \caption{Qualitative comparison of our method (top) and ZSP (bottom) at category-level 3D pose prediction in the wild on samples from PASCAL3D+ and ObjectNet3D (we randomly selected the samples to demonstrate the diversity of the results). For both methods, we overlay our coarse mesh reconstruction in the predicted 3D pose.}%Semantically, the color code reads as, red equals left, green equals back, and blue equals top.\adam{Cut figure}}
    \label{fig:res_objectnet3d}
\end{figure*}

% PASCAL3D CROSS 7
\begin{table*}
\centering
\begin{tabular}{l|l|rrrrrrr|r}
\hline
       & Method &   bicycle &   bus &   car &   chair &   motorcycle &   couch &    tv &   Mean \\
\hline
\multirow{2}{*}{Supervised}  & StarMap \cite{zhou2018starmap} & 83.2 & 94.4 & 90.0 & 75.4 & 68.8 & 79.8 &  85.8 & 82.49 \\
             & VoGE \cite{wang2022voge} & 82.6 & 98.1 & 99.0 & 90.5 & 87.5 & 94.9 & 83.9 & 90.93 \\
\midrule
 \multirow{4}{*}{Unsupervised}            & ZSP        &     \textbf{61.7} & 21.4 & 61.6 &   42.6 &        43.1 &   52.9 & 39.0 &      46.0 \\
         &        &    $\pm$ 14.7 & $\pm$ 8.1 & $\pm$ 11.4 &  $\pm$ 11.3 &      $\pm$  22.1 &  $\pm$ 16.0 & $\pm$ 36.2 &     $\pm$ 17.1 \\

 & Ours &     58.4 & \textbf{79.3} & \textbf{98.2} &   \textbf{51.9} &        \textbf{67.0} &   \textbf{76.6} & \textbf{53.1} &      \textbf{69.2} \\
 &      &      $\pm$ 5.0 &  $\pm$ 11.8 &  $\pm$ 1.0 &   $\pm$ 10.5 &        $\pm$  8.9 &   $\pm$ 13.7 &  $\pm$ 20.4 &     $\pm$  10.2 \\
\hline
\end{tabular}
\caption{3D Pose Estimation in-the-wild on 7 categories of PASCAL3D+. Top two rows show supervised methods (as upper bound) while bottom two rows show unsupervised methods. The reported metric is $30^\circ$ accuracy. 
The mean is averaged over all 7 categories. 
Our method shows superior performance over ZSP, which requires depth annotations.}

\label{tab:res_pascal3d}
\end{table*}

% OBJECTNET3D CROSS 10
\begin{table*}
\centering
 
\begin{tabular}{l|rrrrrrrrrr|r}
\hline
      &   phone &   m'wave &   b'pack &   bench &   cup &   h'dryer &   laptop &   mouse &   remote &   toaster &    \\
\hline
 ZSP  &       46.4 &       50.5 &      \textbf{23.1} &   50.8 & 33.0 &       \textbf{21.7} &    \textbf{60.5} &   28.8 &    41.6 &     28.8 &        \\
     &       $\pm$ 8.4 &     $\pm$  44.3 &    $\pm$  14.9 &  $\pm$  17.0 & $\pm$ 18.7 &    $\pm$   13.3 &   $\pm$  8.9 &  $\pm$ 10.1 &    $\pm$ 18.1 &    $\pm$  8.16 &      \\

 Ours &       \textbf{54.6} &       \textbf{80.3} &      18.0 &   \textbf{62.1} & \textbf{38.2} &       14.1 &    53.3 &   \textbf{44.7} &    \textbf{54.4} &     \textbf{60.6} &        \\
  &       $\pm$  3.4 &      $\pm$  21.7 &    $\pm$   12.0 &   $\pm$  7.1 &  $\pm$ 21.6 &       $\pm$  5.8 &     $\pm$ 9.7 &   $\pm$  9.9 &    $\pm$  4.4 &    $\pm$   2.2 &    \\
    \hline
          &   toilet &   b'cycle &   bus &   car &   chair &   couch &   k'board &   m'cycle &   suitcase &    tv &   Mean  \\
    \hline
     ZSP       &    \textbf{56.3} &      \textbf{58.6} &  30.5 &  60.3 &    36.8 &    55.5 &       \textbf{46.8} &         50.3 &       \textbf{25.8} & 37.9 &        42.2 \\
             &    $\pm$ 13.9 &     $\pm$ 10.4 & $\pm$ 10.6 &  $\pm$ 9.1 &   $\pm$ 10.6 &   $\pm$ 16.1 &      $\pm$ 14.5 &        $\pm$ 20.7 &      $\pm$ 14.3 & $\pm$ 35.1 &       $\pm$ 15.9 \\
     Ours &     39.6 &      57.8 &  \textbf{78.3} &  \textbf{98.1} &    \textbf{52.2} &    \textbf{76.6} &       26.9 &         \textbf{69.0} &       15.5 & \textbf{53.2} &        \textbf{52.4} \\
       &    $\pm$ 12.7 &      $\pm$ 5.1 & $\pm$ 12.1 &  $\pm$ 0.9 &   $\pm$  9.6 &   $\pm$ 12.2 &       $\pm$ 8.4 &         $\pm$ 9.1 &        $\pm$ 5.5 & $\pm$ 22.0 &        $\pm$ 9.8 \\
\hline

\end{tabular}
\caption{3D Pose Evaluation on 10 categories of ObjectNet3D. The mean is averaged over 20 categories. Metric is the $30^\circ$ accuracy. Despite not requiring any depth information, our method significantly outperforms ZSP. } 
\label{tab:res_obj3d_20}   
\end{table*}

As we are not aware of any unsupervised method learning pose estimation from videos, we compare our pose estimation method against two supervised methods \cite{zhou2018starmap, wang2022voge} and ZSP. We provide ZSP with ten uniformly-distributed images of the same reference video that our method uses. Further, we provide ZSP with depth annotations using the category-level CAD models and pose annotations in the PASCAL3D+ and ObjectNet3D data. Despite our method not requiring any depth information, it outperforms ZSP by a large-margin on both PASCAL3D+, see Table \ref{tab:res_pascal3d}, and on ObjectNet3D, see Table \ref{tab:res_obj3d_20}. Qualitative results are depicted in Figure \ref{fig:res_objectnet3d}. We find that ZSP is highly compute intensive, requiring $10.92$ seconds per sample on average, while our proposed method takes only $0.22$ seconds on average. %All runtimes were measured for a batch size of four. 
% hence being almost $50$ times faster.
%for a batch size of four.
%. Both runtimes were evaluated for a batch size of four. 

\textbf{Categorical discussion.} We observe that our method performs better for categories with only small topology changes and deformations, (e.g. car, microwave, couch) compared
to categories with large intra-class variability (e.g. chair). Further, we recognize, that our method even generalizes well from a toybus to a real bus.
Besides that, we analyze, that categories with less available videos (e.g. remote, TV, toilet) on average achieve lower performance.

% PASCAL3D+ 6233 across 7 categories 
% OBJECTNET3D: 12039 across 20 categories

\begin{figure}
  \centering
   \includegraphics[width=\linewidth]{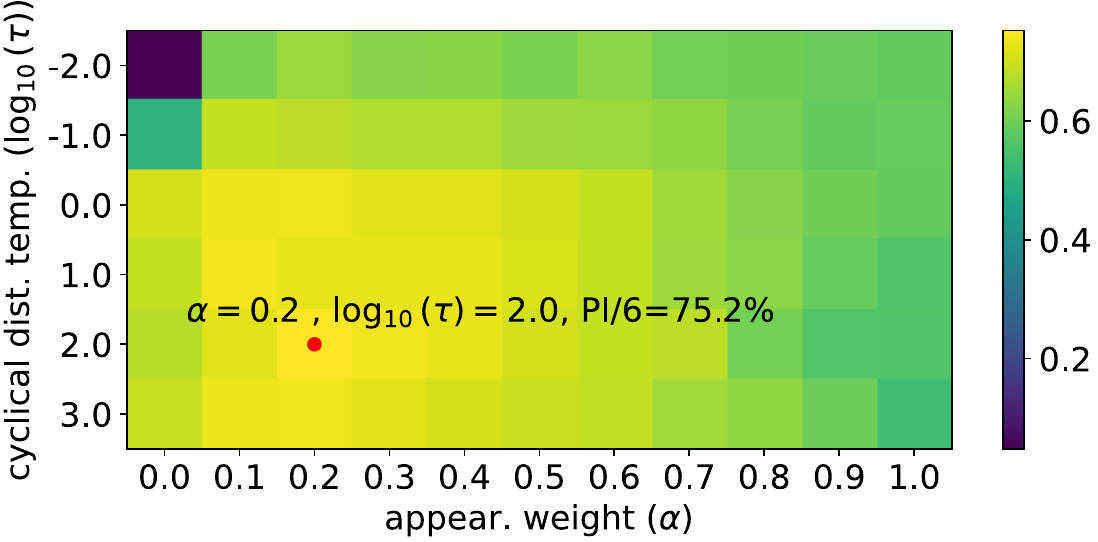}
  \caption{We report the $30^\circ$ accuracy of our alignment method for different choices of our appearance weight $\alpha$ and our cyclical distance temperature $\tau$ resulting in different distances between two meshes with surface features. We see that the maximum accuracy of $75.2\%$ is reached for $\alpha=0.2$ and $\tau=100$. }
    \label{fig:abl_dist}
\end{figure}
\subsection{Ablation}\label{sec:exp:abl}

\textbf{Unsupervised alignment.}
Using the ground-truth annotations of our five references, we measure the effect of both parameters introduced in the alignment method. Namely, the appearance distance weight $\alpha$ and the cyclical distance temperature $\tau$. We remark that for the distance between two meshes with surface features, the appearance weight trades-off feature correspondences versus Euclidean correspondences. Where an appearance weight of $\alpha=0$ means that the distance depends solely on the Euclidean correspondences. Contrarily, an appearance weight of $\alpha=1$ results in solely depending on feature correspondences. Further, our cyclical distance temperature weights each correspondence, implicitly trading-off many low-quality correspondences versus few high-quality ones. Intuitively, increasing the value of $\tau$ results in averaging over more correspondences, while decreasing $\tau$ results in taking only the correspondences with high validity into account. 
In Figure \ref{fig:abl_dist}, we see that both parameters yield a significant impact on the $30^\circ$ accuracy. With an optimum for $\alpha=0.2$ and $\tau=100$. Intuitively, this means that taking many correspondences into account is more beneficial. Additionally, the  Euclidean correspondences are weighted four times as much as the feature correspondences. Besides that, the ablation shows that while many correspondences are essential for using solely Euclidean correspondences, the opposite is true when using solely feature correspondences.

\textbf{3D pose estimation in-the-wild.} Following the alignment, the in-the-wild 3D pose estimation task can also be solved using neural network regression with the 6D rotation representation proposed in \cite{zhou2019continuity}. However, we observe that the results are worse than our 3D template learning method combined with render-and-compare, see Table \ref{tab:abl_model_design}. 

\begin{table}[]
    \centering
    \begin{tabular}{l|l|rr}
    \hline
        & Method & Acc. $30^\circ$  &   Acc. $15^\circ$ \\ \hline
       \multirow{2}{*}{PASCAL3D+} & Regression & 66.0 &   25.8  \\ 
                                  & Ours & \textbf{69.2} & \textbf{41.3}  \\ \hline
       \multirow{2}{*}{ObjectNet3D} & Regression & 44.5 & 16.4  \\ 
                                    & Ours & \textbf{52.4} & \textbf{25.5} \\ \hline
    \end{tabular}
    
    \caption{Average $30^\circ$ and $15^\circ$ accuracies on PASCAL3D+ and ObjectNet3D for using directly neural network regression.}
    \label{tab:abl_model_design}
\end{table}

\subsection{Limitations}
We have proposed a model which substantially outperforms existing applicable baselines for the task of unsupervised category-level 3D pose estimation in-the-wild.
However, our proposed method does not yet reach the performance of fully supervised baselines. 
One advancement we aspire is to relax the rigidity constraint of our shape model. Therefore, we plan to leverage the aligned reconstructions  and introduce a parameterized model for the shape. A deformable shape would yield the potential to improve the correspondence learning as well as the subsequent matching of features at inference. Moreover, we see a future research direction in enabling the model to learn from a continuous stream of data, instead of building on a set of pre-recorded videos. This would even better reflect the complex real-world scenarios of embodied agents.
%

% well-suited: categories must be in CO3D, and there must be comparison available

% ZSP categories: 
% well-suited: bicycle, hydrant, motorcycle, teddy, (toytruck, toytrain, car, toybus, keyboard, handbag, remote, toyplane, toilet, hairdryer, mouse, toaster, chair, laptop, book, backpack)

% PASCAL3D categories: 
% well-suited: (7) car, motorcycle, bicycle, chair, (sofa, tv, bottle)
% excluded due to toys: aeroplane, train, bus, 
% not available in CO3D: dining table, boat

% ObjectNet3D categories
% nemo and starmap report on :
% bed, bookshelf, calculator, cellphone, laptop, cabinet, guitar, iron, knife, microwave, pen, pot, rifle, slipper, stove, toilet, tub, wheelchair

% in CO3D: backpack, bench, bicycle, bottle, car, cellphone, chair, sofa, cup, hairdryer, keyboard, laptop, microwave, motorcycle, mouse, suitcase, toaster, toilet, (bus, aeroplane, train), tv
% skateboard, sign

% evaluated by starmap, pose contrast, (18)
%bed, bookshelf, calculator, cellphone, computer, filing cabinet, guitar, iron, knife, microwave, pen, %pot, rifle, slipper, stove, toilet, tub, wheelchair

% well-suited: (3-4) cellphone, laptop, microwave, (toilet), (all others in CO3D)

% additional (15): backpack, bench, bicycle, bottle, car, chair, sofa, cup, hairdryer, keyboard, motorcycle, mouse, suitcase, toaster, tv

% total well-suited: (17-18)

% additonal zsp (9): hydrant, teddy, toytruck, toytrain, toybus, handbag, remote, toyplane, book
% missing zsp (7): cellphone, bench, bottle, sofa, cup, suitcase, tv, microwave

% pix3d Bed Bookcase Chair Desk Misc Sofa Table Tool Wardrobe: only chair, sofa, bookcase

%% file: sec/5_conclusion.tex
\section{Conclusion}
In this paper, we have proposed a highly challenging (but realistic) task: unsupervised category-level 3D pose estimation from object-centric videos. 
In our proposed task, a model is required to align object-centric videos of instances of an object category without having any pose-labelled data.
Subsequently, the model learns a 3D representation from the aligned videos to perform 3D category-level pose estimation in the wild.
Our task defines a complex real-world problem which requires both semantic and geometric understanding of objects, and we demonstrate that existing baselines cannot solve the task. 
We further proposed a novel method for unsupervised learning of category-level 3D pose estimation that follows a two-step process: 
1) A multi-view alignment procedure that determines canonical camera poses across videos with a novel and robust cyclic distance formulation for geometric and appearance matching.
2) Learning dense correspondences between images and a prototypical 3D template by predicting, for each pixel in a 2D image, a feature vector of the corresponding vertex in the template mesh.
The results showed that our proposed method achieves large improvements over all baselines, and we hope that our work will pave the ground for future advances in this important research direction.

\section{Acknowledgement}

Adam Kortylewski acknowledges support for his Emmy Noether Research Group funded by the German Science Foundation (DFG) under Grant No. 468670075.

%% file: sec/6_supplementary.tex
\appendix
\section{Supplementary}

\subsection{Alignment Ablation}

\begin{table*}[]
\centering
\begin{tabular}{l|c|c|c|rrr}
\hline
Feat. Encoder & Avg. Feat. & Vertex Feat. Dist. & Refine & \multicolumn{1}{l}{Acc. $30^\circ$} & \multicolumn{1}{l}{Acc. $15^\circ$} & \multicolumn{1}{l}{Acc. $10^\circ$} \\ \hline
DINOv2 ViT-S/14      & \checkmark      & min-min         &            & 26.0                          & 22.6                        & 21.2                        \\
DINOv2 ViT-S/14      &            & min-min         &            & 77.0                        & 61.6                        & 46.1                        \\
DINOv2 ViT-S/14      &            & mean-min    &            & 79.1                        & 62.6                        & 47.7                        \\
DINOv2 ViT-S/14      &            & mean-min    & \checkmark          & 79.9                        & 69.7                        & 56.9                        \\
DINOv2 ViT-B/14      &            & mean-min    & \checkmark          & 84.3                        & 73.4                        & 60.6                        \\
DINOv1 ViT-S/8       &            & mean-min    & \checkmark          & 74.4                        & 63.0                          & 51.8                        \\ \hline
\end{tabular}
\caption{Ablation for 7D alignment on the CO3D dataset, averaged across 20 categories. }
\label{tab:abl_zsp_20}
\end{table*}

In Table \ref{tab:abl_zsp_20}, we study the effect for different distance calculations between two vertices containing many features from many viewpoints. We observe, that averaging over the features in a single vertex before calculating the distance to another vertex reduces the performance drastically. Further, calculating the distance by averaging over the bi-directional nearest neighbor distances, slightly improves the performance compared to taking the minimum distance over the bi-directional nearest neighbor distances. 

Besides that, we show that refining the initial alignment using few gradient-based optimization steps improves the results, especially with respect to the more fine-grained $10^\circ$ and $15^\circ$ accuracies. 

In the same table, we observe the significant effect for using different feature extractors.

\subsection{In-the-Wild 3D Pose Estimation Ablation}

In Table \ref{tab:abl_pose3d}, we ablate our 3D pose estimation method for various amount of training data. We show the results for maximum 5, 10, 20, or 50 videos per category.

\begin{table*}[]
    \centering
    \begin{tabular}{c|c|rrr}    
    \hline
        & \# Videos	   & Acc. $30^\circ$  &   Acc. $15^\circ$ &   Acc. $10^\circ$  \\ \hline
        \multirow{5}{*}{PASCAL3D+} & 5 & 56.3 & 33.2 & 21.6 \\
                                   & 10 & 58.2 & 33.4 & 21.2 \\
                                   & 20 & 65.2 & 38.9 & 25.0   \\
                                   & 50 & 69.2 & 41.3 & 25.5 \\
          \hline
       \multirow{5}{*}{ObjectNet3D} & 5 & 45.3 & 21.0   & 11.7 \\
                                    & 10 & 45.8 & 20.9 & 11.4 \\
                                    & 20 & 49.1 & 24.1 & 13.8 \\
                                    & 50 & 52.4 & 25.5 & 14.1 \\
                                     \hline
    \end{tabular}
    \caption{Ablation for 3D pose estimation on PASCAL3D+ and ObjectNet3D.}
    \label{tab:abl_pose3d}
\end{table*}

\subsection{Mesh Reconstruction from Videos}

\begin{figure*}
    \centering
    \includegraphics[width=\linewidth]{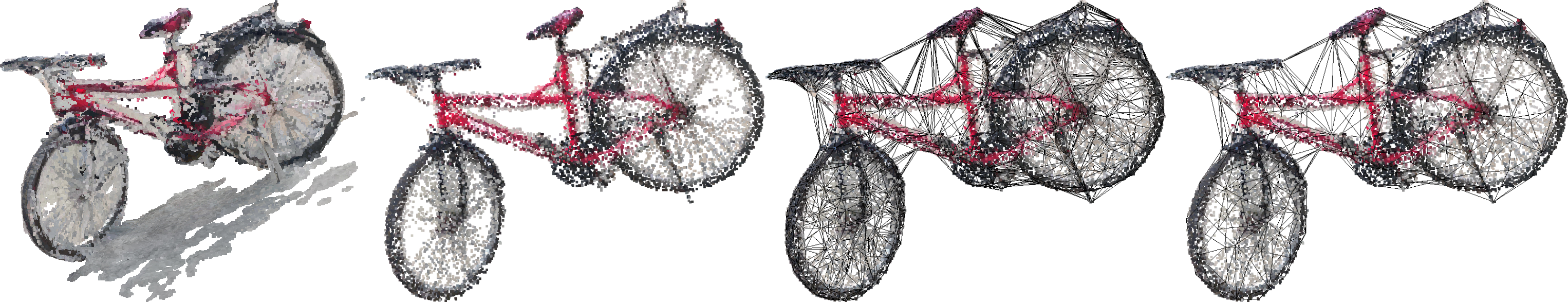}
    \caption{Steps of estimating a coarse mesh given a noisy point cloud (left). First, we clean the point cloud using the object segmentation for each frame. Second, we estimate a coarse mesh using alpha shapes \cite{kirkpatrick1983shape}. Third, we remove faces using quadratic decimation \cite{garland1997surface}.}
    \label{fig:mesh-extraction}
\end{figure*}

Using structure-from-motion \cite{schonberger2016colmap} we obtain a point cloud $P=\{ \vi \in \mathbb{R}^3 \}$ for each video. Further, we reconstruct a coarse mesh using three steps. First, we randomly downsample the point cloud to 20000 points and clean it using the object segmentation provided by CO3D. Therefore, we compute an average ratio of visibility for each point $\vi$ by projecting it in all $N$ frames, using the respective projection $\pi_j$ for frame $j$, and averaging over the respective visibilities $\delta_j(\pi_j(\vi))$ as follows
\begin{equation}
        p(\vi) = \frac{1}{N} \sum_{j=1}^N \delta_j(\pi_j(\vi)).
\end{equation}
Hereby we set $\delta_j(\pi_j(\vi))=0$, if the projected vertex is not inside the frame. We filter out all points which ratio of visibility lies below 60\%.
Second, we use alpha shapes \cite{kirkpatrick1983shape} to estimate a coarse shape from the clean point cloud. Figuratively speaking, this algorithm starts off with a convex volume and then iteratively carves out spheres while preserving all original points. We set the size of the sphere to 10 times the particle size, where the particle size is the average distance of each point to its 5th closest point. Third, we use quadratic mesh decimation \cite{garland1997surface} to end up with a maximum of 500 faces. This method iteratively contracts a pair of vertices, minimizing the projective error with the faces normals. All steps are visualized in Figure \ref{fig:mesh-extraction}.

\subsection{Videos Filtering}

\begin{figure}
    \centering
    \includegraphics[width=\linewidth]{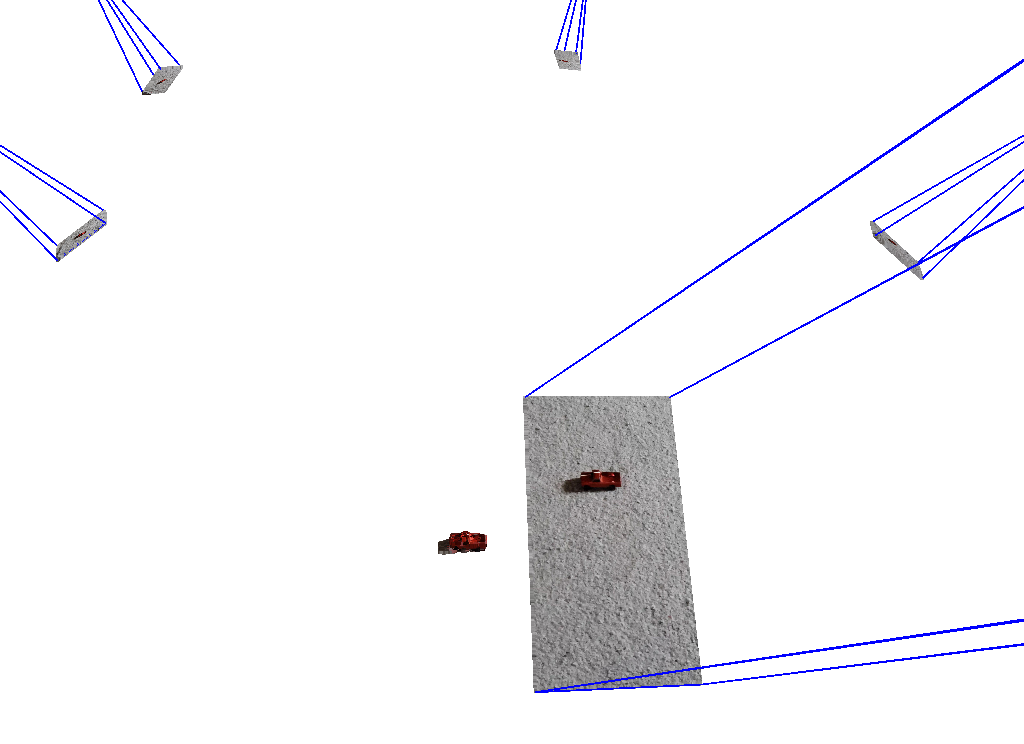}
    \caption{Rejected video type a), object is too far away from the camera. }
    \label{fig:no-mask-coverage}
\end{figure}

\begin{figure}
    \centering
    \includegraphics[width=\linewidth]{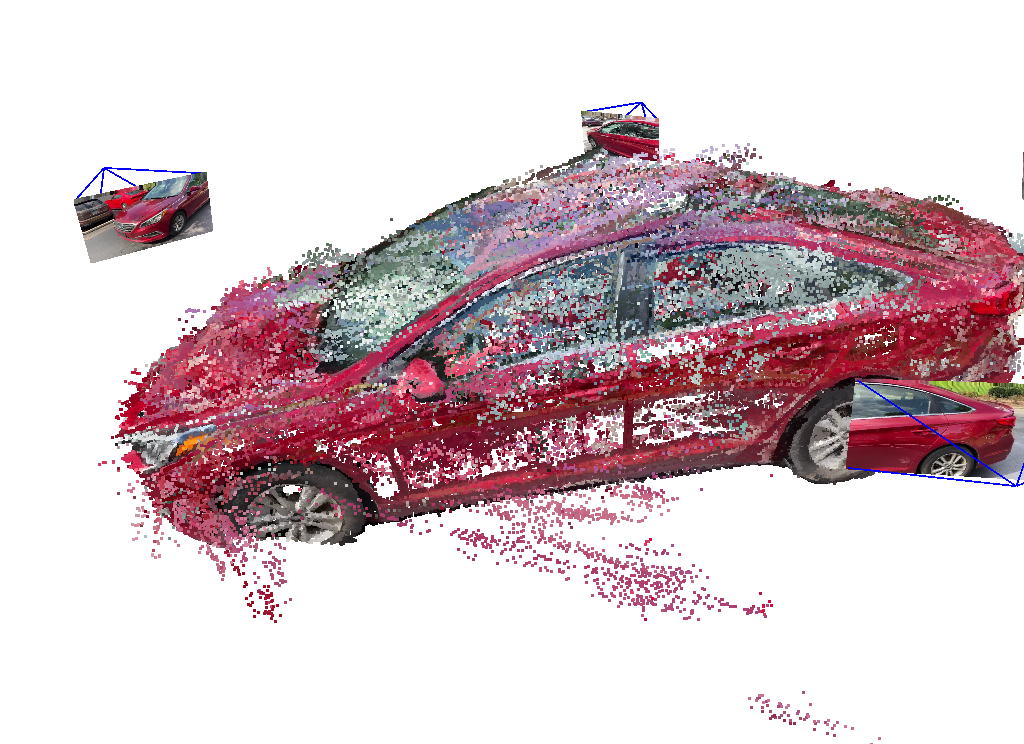}
    \caption{Rejected video type b), object is too close to the camera. }
    \label{fig:not-centered}
\end{figure}

\begin{figure}
    \centering
    \includegraphics[width=\linewidth]{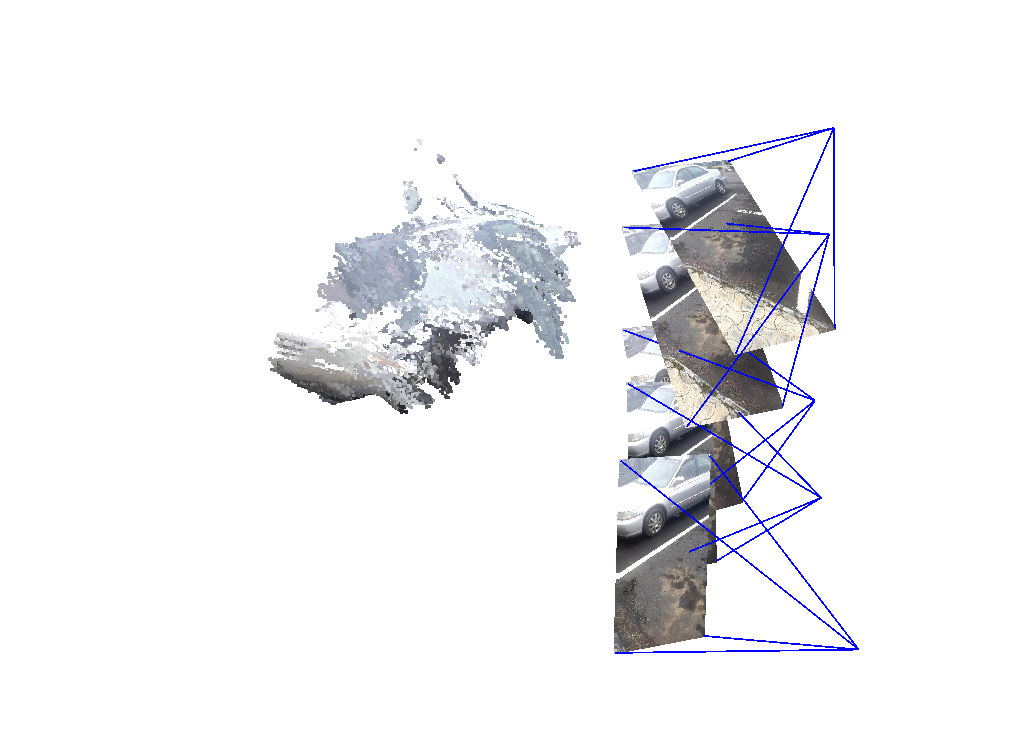}
    \caption{Rejected video type c), the variance of viewpoints is too small.}
    \label{fig:no-viewpoints-variance}
\end{figure}

We filter out three types of videos. Type a), object is too far away from the camera. Type b), object is too close to the camera. Type c), the variance of viewpoints is too small. Type a) is not ideal because the point cloud and the images yield only few details of the object. Type b) is problematic because the close-ups prevent us from robustly cleaning the noisy point cloud as there is less information accumulated from the object segmentations. Type c) results in a very noisy or even broken structure-from-motion. 
For the identification of type a), object is too far away from the camera, we use the average object visibility $\bar{\delta}$ over all $N$ frames width $U$ and height $V$ formally defined as
\begin{equation}
    \begin{aligned}
        \bar{\delta} = \frac{1}{N U V} \sum_{j=1}^N \sum_{u=1}^U \sum_{v=1}^V \delta_j(u, v).
    \end{aligned}
\end{equation}
We require an average object visibility of at least 10\%. A filtered out video is illustrated in Figure \ref{fig:no-mask-coverage}.
For the identification of type b), the object is too close to the camera, we use the projection of the 3D center into all frames, expecting it to be in the center of the frames. We compute the 3D center $c$ using the camera rays with position $r_j$ and direction $n_j$ by minimizing its projected distances to the rays
\begin{equation}
    \begin{aligned}
        \min_c \sum_j \langle (c - r_j)^T n_j, (c - r_j)^T n_j \rangle.
    \end{aligned}
\end{equation}
It can be shown that this resolves to the following system of linear equations
\begin{equation}
    \begin{aligned}
        \left(\sum_j n_j n_j^T\right) x = \sum_j n_j n_j^T r_j.
    \end{aligned}
\end{equation}
With the outer product $n_j n_j^T \in \mathbb{R}^{3\times3}$. For a correct camera focus, we expect the projected 3D center to lie within the centered rectangle spanning 60\% of the image width and height. In total, we require 80\% of the frames to be focused on the 3D center. A negative example is provided in Figure \ref{fig:no-mask-coverage}. A filtered out video is illustrated in Figure \ref{fig:not-centered}.

For the identification of type c), the variance of viewpoints is too small, we subtract the center $c$ of all camera positions $r_j$ and normalize them to lie on the unit sphere. Further, we divide the unit sphere into 38 bins and calculate the viewpoint coverage as percentage of viewpoint bins covered. We require a viewpoint coverage for each video of 15\%. A rejected video is shown in Figure \ref{fig:no-viewpoints-variance}.
% 5 steps each dimension [-1, 1], offset to norm=1 < 1/5, 38 viewpoints coverage 0.15 -> 6 different viewpoints from 38 bins

\subsection{ObjectNet3D}

\begin{figure*}
    \centering
    \includegraphics[width=\linewidth]{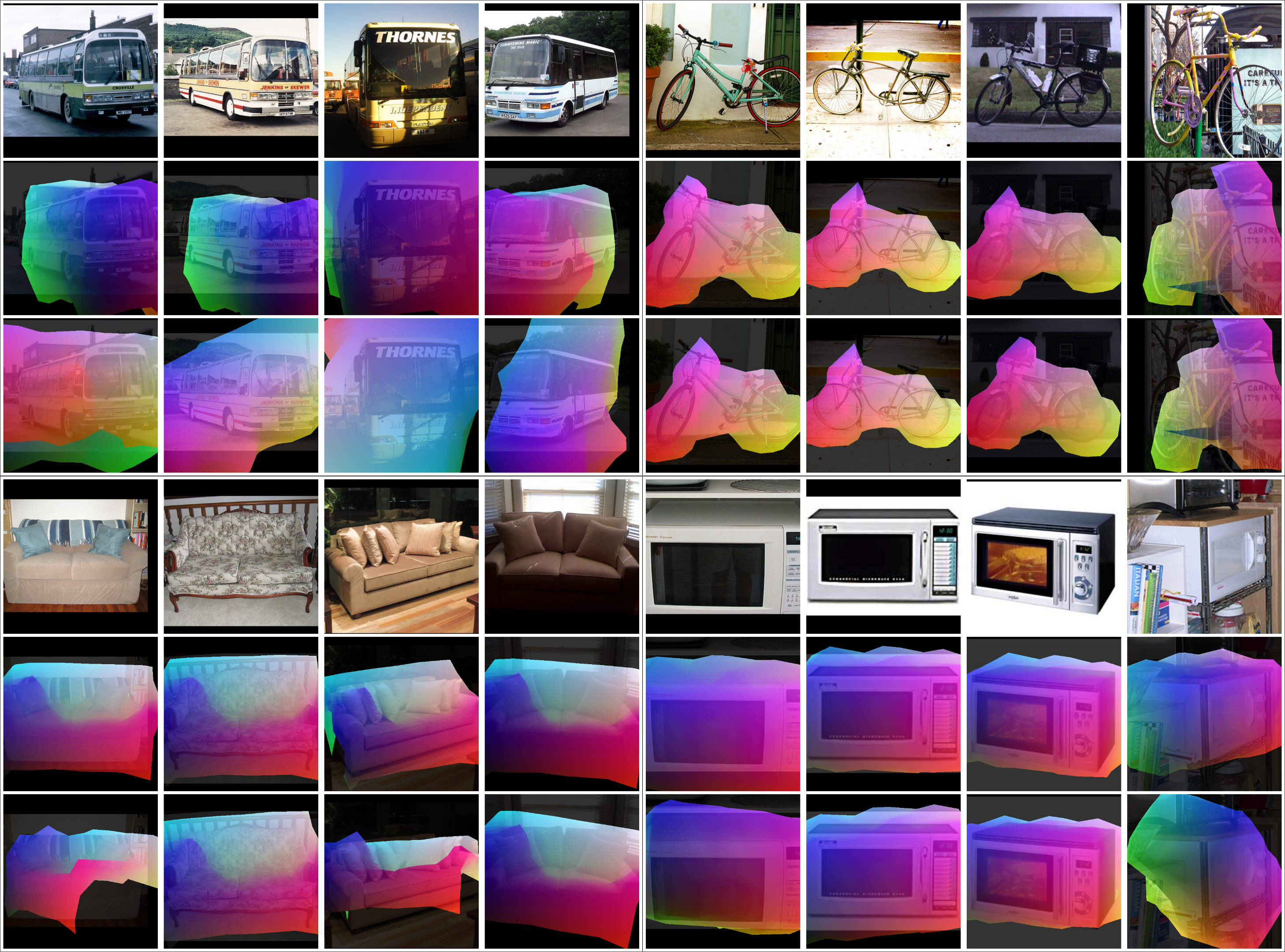}
    \caption{Qualitative comparison of our method (top) and ZSP (bottom) at category-level 3D pose prediction in the wild on samples from ObjectNet3D (we randomly selected the samples to demonstrate the diversity of the results). For both methods, we overlay our coarse mesh reconstruction in the predicted 3D pose.}
    \label{fig:res_pose_in_the_wild_more}
\end{figure*}

We report more qualitative results are visualized in Figure \ref{fig:res_pose_in_the_wild_more}.

\begin{figure*}
    \centering
    \includegraphics[width=\linewidth]{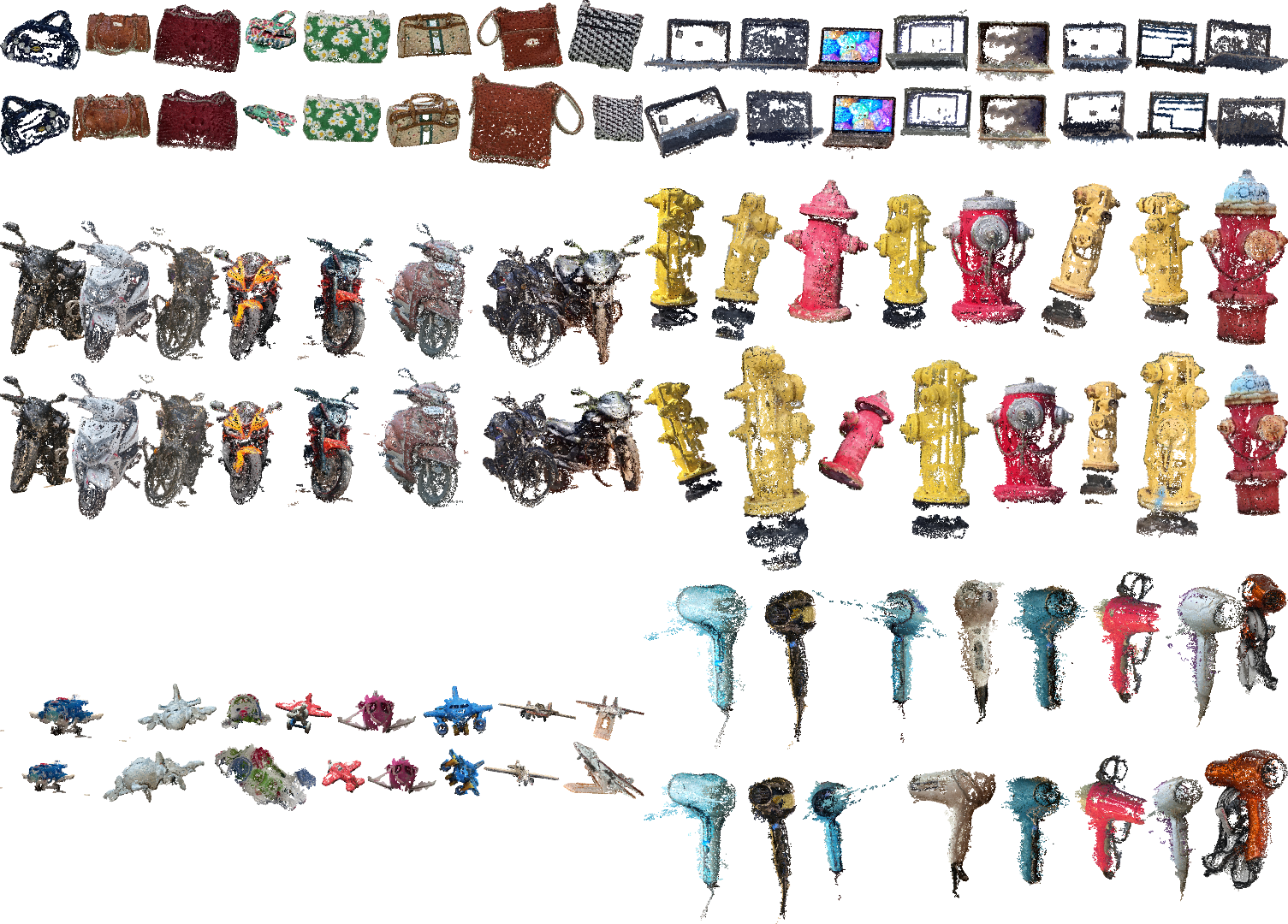}
    \caption{Qualitative comparison of two unsupervised alignment methods. The first row illustrates the alignment of our proposed method. The second row shows the alignment using ZSP \cite{goodwin2022zsp}. For both methods, we utilize the 5th object instance from the left as a reference point. Our proposed method proves to be more precise than ZSP. }
    \label{fig:res_alignment_more}
\end{figure*}